%% file: lrec2020W-xample-kc.tex
\documentclass[10pt, a4paper]{article}
\usepackage{lrec}
\usepackage{multibib}
\newcites{languageresource}{Language Resources}
\usepackage{graphicx}
\usepackage{tabularx}
\usepackage{soul}
\usepackage{titlesec}
\titleformat{\section}{\normalfont\large\bf\center}{\thesection.}{1em}{}
\titleformat{\subsection}{\normalfont\SmallTitleFont\bf\raggedright}{\thesubsection.}{1em}{}
\titleformat{\subsubsection}{\normalfont\normalsize\bf\raggedright}{\thesubsubsection.}{1em}{}
\renewcommand\thesection{\arabic{section}}
\renewcommand\thesubsection{\thesection.\arabic{subsection}}
\renewcommand\thesubsubsection{\thesubsection.\arabic{subsubsection}}

\usepackage{epstopdf}
\usepackage[utf8]{inputenc}

\usepackage{hyperref}
\usepackage{xstring}

\usepackage{color}

\title{Gender Representation in Open Source Speech Resources}

\name{Mahault Garnerin, Solange Rossato, Laurent Besacier}

\address{LIG, Univ. Grenoble Alpes, CNRS, Grenoble INP, FR-38000 Grenoble, France\\
         firstname.lastname@univ-grenoble-alpes.fr\\}

\abstract{
With the rise of artificial intelligence (AI) and the growing use of deep-learning architectures, the question of ethics, transparency and fairness of AI systems has become a central concern within the research community. We address transparency and fairness in spoken language systems by proposing a study about  gender representation in speech resources available through the Open Speech and Language Resource platform. We show that finding gender information in open source corpora is not straightforward and that gender balance depends on other corpus characteristics (elicited/non elicited speech, low/high resource language, speech task targeted). The paper ends with recommendations about metadata and gender information for researchers in order to assure better transparency of the speech systems built using such corpora. \\ \newline  }

\begin{document}

\maketitleabstract

\section{Introduction}

The ever growing use of machine learning has put data at the center of the industrial and research spheres. Indeed, for a system to learn how to associate an input X to an output Y, many paired examples are needed to learn this mapping process. This need for data coupled with the improvement in computing power and algorithm efficiency has led to the era of big data. But data is not only needed in mass, but also with a certain level of quality. In this paper we argue that one of the main quality of data is its \emph{transparency}.

In recent years, concerns have been raised about the biases existing in the systems. A well-known case in Natural Language Processing (NLP) is the example of word embeddings, with the studies of \newcite{bolukbasi2016man} and \newcite{caliskan2017semantics} which showed that data are socially constructed and hence encapsulate a handful of social representations and power structures, such as gender stereotypes. Gender-bias has also been found in machine translation tasks \cite{vanmassenhove2018getting}, as well as facial recognition \cite{buolamwini2018gender} and is now at the center of research debates. In previous work, we investigated the impact of gender imbalance in training data on the performance of an automatic speech recognition (ASR) system, showing that the under-representation of women led to a performance bias of the system for female speakers \cite{Garnerin:2019:GRF:3347449.3357480}. 

In this paper, we survey the gender representation within an open platform gathering speech and language resources to develop speech processing tools. The aim of this survey is twofold: firstly, we investigate the gender balance within speech corpora in terms of speaker representation but also in terms of speech time available for each gender category. Secondly we propose a reflection about general practices when releasing resources, basing ourselves on some recommendations from previous work.

\textbf{Contributions.} The contributions of our work are the following:
\begin{itemize}
    \item an exploration of 66 different speech corpora in terms of gender, showing that gender balance is achieved in terms of speakers in elicited corpora, but that it is not the case for non-elicited speech, nor for the speech time allocated to each gender category
    \item  an assessment of the global lack of meta-data within free open source corpora, alongside recommendations and guidelines for resources descriptions, based on previous work
\end{itemize}

\section{OpenSLR}

Open Speech Language Resources\footnote{\url{http://www.openslr.org.}} (OpenSLR) is a platform created by Daniel Povey. It provides a central hub to gather open speech and language resources, allowing them to be accessed and downloaded freely. OpenSLR currently\footnote{Last checked on November 14th, 2019.} hosts 83 resources. These resources consist of speech recordings with transcriptions but also of softwares as well as lexicons and textual data for language modeling. As resources are costly to produce, they are most of the time a paying service. Therefore it is hard to study gender representation at scale.  We thus focus on the corpora available on OpenSLR due to their free access and to the fact that OpenSLR is explicitly made to help develop speech systems (mostly ASR but also text-to-speech (TTS) systems). In our work, we focus on speech data only.

Out of the 83 resources gathered on the platform, we recorded 53 speech resources. We did not take into account multiple releases of the same corpora but only kept the last version (e.g.\ TED LIUM \cite{hernandez2018ted}) and we also removed subsets of bigger corpora (e.g.\ LibriTTS corpus \cite{zen2019libritts}). We make the distinction between a resource and a corpus, as each resource can contain several languages (e.g.\ Vystadial \citelanguageresource{korvas_2014}) or several accent/dialect of a same language (e.g.\ the crowdsourced high-quality UK and Ireland English Dialect speech data set \citelanguageresource{google_uken_2019}). In our terminology, we define a corpus as monolingual and monodialectal, so resources containing different dialects or languages will be considered as containing different corpora.

We ended up with 66 corpora, in 33 different languages with 51 dialect/accent variations. The variety is also great in terms of speech types (elicited and read speech, broadcast news, TEDTalks, meetings, phonecalls, audiobooks, etc.), which is not suprising, given the many different actors who contributed to this platform. We consider this sample to be of reasonable size to tackle the question of gender representation in speech corpora.\footnote{Our case study does not claim to be exhaustive and future investigations should definitely include data sets provided by resource agencies such as ELRA or LDC.} OpenSLR also constitutes a good indicator of general practice as it does not expect a defined format nor does have explicit requirements about data structures, hence attesting of what metadata resources creators consider important to share when releasing resources for free on the Web. 

\section{Methodology}

In order to study gender representation within speech resources, let us start by defining what gender is. In this work, we consider gender as a binary category (male and female speakers). Nevertheless, we are aware that gender as an identity also exists outside of these two categories, but we did not find any mention of non-binary speakers within the corpora surveyed in our study.

Following work by \newcite{doukhan2018open}, we wanted to explore the corpora looking at the number of speakers of each gender category as well as their speech duration, considering both variables as good features to account for gender representation. After the download, we manually extracted information about gender representation in each corpus.

\subsection{Speaker Information and Lack of Meta-Data}
 
The first difficulty we came across was the general absence of information. As gender in technology is a relatively recent research interest, most of the time gender demographics are not made available by the resources creators. So, on top of the further-mentioned general corpus characteristics (see Section~\ref{corpus_char}), we also report in our final table where the gender information was found and whether it was provided in the first place or not. 

The \emph{provided} attribute corresponds to whether gender info was given somewhere, and the \emph{found\_in} attribute corresponds to where we extracted the gender demographics from. The different modalities are \emph{paper}, if a paper was explicitly cited along the resource, \emph{metadata} if a metadata file was included, \emph{indexed} if the gender was explicitly indexed within data or if data was structured in terms of gender and \emph{manually} if the gender information are the results of a manual research made by ourselves, trying to either find a paper describing the resources, or by relying on regularities that seems like speaker ID and listening to the recordings. We acknowledge that this last method has some methodological shortcomings: we relied on our perceptual stereotypes to distinguish male from female speakers, most of the time for languages we have no knowledge of, but considering the global lack of data, we used it when corpora were small enough in order to increase our sample size.

\subsection{Speech Time Information and Data Consistency} 

The second difficulty regards the fact that speech time information are not standardised, making impossible to obtain speech time for individual speakers or gender categories. When speech time information is provided, the statistics given do not all refer to the same measurements. Some authors report speech duration in hours {e.g.\ \citelanguageresource{panayotov2015librispeech,hernandez2018ted}}, some the number of utterances (e.g \cite{Juan2015}) or sentences (e.g.\ \citelanguageresource{google_uken_2019}), the definition of these two terms never being clearly defined. We gathered all information available, meaning that our final table contains some empty cells, and we found that there was no consistency between speech duration and number of utterances, excluding the possibility to approximate one by the other. As a result, we decided to rely on the size of the corpora as a (rough) approximation of the amount of speech data available, the text files representing a small proportion of the resources size. This method however has drawbacks as not all corpora used the same file format, nor the same sampling rate. Sampling rate has been provided as well in the final table, but we decided to rely on qualitative categories, a corpus being considered small if its size is under 5GB, medium if it is between 5 and 50GB and large if above.\footnote{A reviewer rightly pointed out that we could estimate speech duration having its file size, sampling rate and number of bits for quantification, but due to the difficulty to gather all these information and the variety of resources structures, we left it as future work perspective}

\subsection{Corpora Characteristics}
\label{corpus_char}

The final result consists of a table\footnote{The final table and the script used for the analysis are available at: \url{https://github.com/mgarnerin/openslr\_gender\_survey}.} reporting all the characteristics of the corpora. The chosen features are the following: 

\begin{itemize}
\item the resource identifier (\emph{id}) as defined on OpenSLR
\item the language (\emph{lang})
\item the dialect or accent if specified (\emph{dial})
\item the total number of speakers as well as the number of male and female speakers (\emph{\#spk}, \emph{\#spk\_m}, \emph{\#spk\_f})
\item the total number of utterances as well as the total number of utterances for male and female speakers (\emph{\#utt}, \emph{\#utt\_m}, \emph{\#utt\_f}) 
\item the total duration, or speech time, as well as the duration for male and female speakers (\emph{dur}, \emph{dur\_m}, \emph{dur\_f}) 
\item the size of the resource in gigabytes (\emph{sizeGB}) as well as a qualitative label (\emph{size}, taking its value between ``big", ``medium", ``small")
\item the sampling rate (\emph{sampling})
\item the speech task targeted for the resource (\emph{task})
\item is it \emph{elicited} speech or not: we define as non-elicited speech data which would have existed without the creation of the resources (e.g\ TedTalks, audiobooks, etc.), other speech data are considered as elicited 
\item the language status (\emph{lang\_status}): a language is considered either as high- or low-resourced. The language status is defined from a technological point of view (i.e.\ are there resources or NLP systems available for this language?). 
It is fixed at the language granularity (hence the name), regardless of the dialect or accent (if provided).
\item the year of the release (\emph{year})
\item the  authors of the resource (\emph{producer})
\end{itemize}

\section{Analysis}

\input{tables/info_prov.tex}
\input{tables/prov_spk_dur.tex}

\subsection{Gender Information Availability}

Before diving into the gender analysis, we report the number of corpora for which gender information was provided. Indeed, 36.4\% of the corpora do not give any gender information regarding the speakers. Moreover, almost 20\% of the corpora do not provide any speaker information whatsoever. 
Table~\ref{tab:info_prov} sums up the number of corpora for which speaker's gender information was provided and if it was, where it was found. We first looked at the metadata file if available. If no metadata was provided, we searched whether gender was indexed within the data structure. At last, if we still could not find anything, we looked for a paper describing the data set. This search pipeline results in ordered levels for our \emph{found\_in} category, meaning papers might also be available for corpora with the ``metadata" or ``indexed" modalities. 

When gender information was given it was most of the time in terms of number of speakers in each gender categories, as only five corpora provide speech time for each category. Table \ref{tab:info_prov_gender} reports  what type of information was provided in terms of gender, in the subset of the 42 corpora containing gender information. We observe that gender information is easier to find when it regards the number of speakers, than when it accounts for the quantity of data available for each gender group. Due to this lack of data, we did not study the speech time per gender category as intended, but we relied on utterance count when available. It is worth noticing however, that we did not find any consistency between speech time and number of utterances, so such results must be taken with caution.

Out of the 42 corpora providing gender information, 41 reported speaker counts for each gender category. We manually gathered speaker gender information for 7 more corpora, as explained in the previous section, reaching a final sample size of 47 corpora.\footnote{The Free ST Chinese Mandarin Corpus \citelanguageresource{free_chinese} provided gender information, but we did not manage to use it, hence a total of 47 and not 48.}

\subsection{Gender Distribution Among Speakers}
\subsubsection{Elicited vs Non-Elicited Data}

Generally, when gender demographics are provided, we observe the following distribution: out of the 6,072 speakers, 3,050 are women and 3,022 are men, so parity is almost achieved. We then look at whether data was elicited or not, non-elicited speech being speech that would have existed without the corpus creation such as TEDTalks, interviews, radio broadcast and so on. We assume that if data was not elicited, gender imbalance might emerge. Indeed, non-elicited data often comes from the media, and it has been shown, that women are under-represented in this type of data \cite{GMMP2015}. This disparity of gender representation in French media \cite{CSA2017,doukhan2018open} precisely led us to the present survey. Our expectations are reinforced by examples such as the resource of Spanish TEDTalks, which states in its description regarding the speakers that \emph{``most of them are men"} \citelanguageresource{mena_2019}. We report results in Table~\ref{tab:gender_rep_el}.

In both cases (respectively elicited and non-elicited speech), gender difference is relatively small (respectively 5.6 percentage points and 5.8 points), far from the 30 percentage points difference observed in \cite{Garnerin:2019:GRF:3347449.3357480}. A possible explanation is that either elicited or not, corpora are the result of a controlled process, so gender disparity will be reduced as much as possible by the corpus authors. However, we notice that, apart from Librispeech \cite{panayotov2015librispeech}, all the non-elicited corpora are small corpora. When removing Librispeech from the analysis, we observe a 1/3-2/3 female to male ratio, coherent with our previous findings. This can be explained by the care put by the creators of the Librispeech data set to \emph{"[ensure] a gender balance at the speaker level and in terms of
the amount of data available for each gender"} \cite{panayotov2015librispeech}, while general gender disparity is observed in smaller corpora.

What emerges from these results is that when data sets are not elicited or carefully balanced, gender disparity creeps in. This gender imbalance is not observed at the scale of the entire OpenSLR platform, due to the fact that most of the corpora are elicited (89.1\%). Hence, the existence of such gender gap is prevented by a careful control during the data set creation process.

\input{tables/gender_elicited.tex}

\subsubsection{High-resource vs Low-resource Languages}

In the elicited corpora made available on OpenSLR, some are of low-resource languages other high-resource languages (mostly regional variation of high-resources languages). When looking at gender in these elicited corpora, we do not observe a difference depending on the language status. However, we can notice that high-resource corpora contain twice as many speakers, all low-resource language corpora being small corpora.


\input{tables/gender_lang_status.tex}

\subsubsection{``How Can I Help?": Spoken Language Tasks}

Speech corpora are built in order to train systems, most of the time ASR or TTS ones. We carry out our gender analysis taking into account the task addressed and obtain the results reported in Table~\ref{tab:gender_rep_task}. We observe that if gender representation is almost balanced within ASR corpora, women are better represented in TTS-oriented data sets. This can be related to the UN report of recommendation for gender-equal digital education stating that nowadays, most of the vocal assistants are given female voices which raises educational and societal problems \cite{west2019d}. This gendered design of vocal assistants is sometimes justified by relying on gender stereotypes such as ``female voices are perceived as more helpful, sympathetic or pleasant." TTS systems being often used to create such assistants, we can assume that using female voices has become general practice to ensure the adoption of the system by the users. This claim can however be nuanced by \newcite{nass2005wired} who showed that other factors might be worth taking into account to design gendered voices, such as social identification and cultural gender stereotypes.

\input{tables/asr_tts_table.tex}
\input{tables/women_tts_table.tex}


\subsection{Speech Time and Gender}
Due to a global lack of speech time information, we did not analyse the amount of data available per speaker category. However, utterance counts were often reported, or easily found within the corpora. We gathered utterance counts for a total of 32 corpora. We observe that if gender balance is almost achieved in terms of number of speakers, at the utterance level, men speech is more represented. But this disparity is only the effect of three corpora containing 51,463 and 26,567 \citelanguageresource{korvas_2014} and 8376 \citelanguageresource{mena_2019} utterances for male speakers, while the mean number of utterances per corpora is respectively 1942 for male speakers and 1983 for female speakers. Removing these three outliers, we observe that utterances count is balanced between gender categories.

It is worth noticing, that the high amount of utterances of the outliers is surprising considering that these three corpora are small (2.1GB, 2.8GB) and medium (5.2GB). This highlights the problem of the notion of utterance which is never being explicitly defined. Such difference in granularity is thus preventing comparison between corpora.

\subsection{Evolution over Time}

\begin{figure}[!h]
\begin{center}
\includegraphics[scale=0.4]{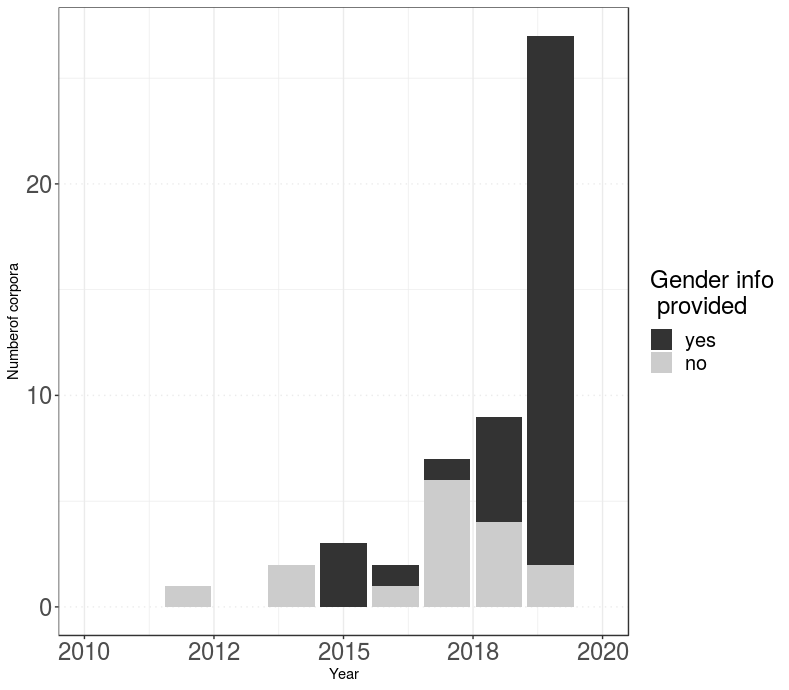} 
\caption{Evolution of gender information availability in OpenSLR resources from 2010 to 2019.}
\label{fig.1}
\end{center}
\end{figure}

When collecting data, we noticed that the more recent the resources, the easier it was to find gender information, attesting of the emergence of gender in technology as a relevant topic. As pointed out by Kate \newcite{crawford2017nips} in her NeurIPS keynote talk, fairness in AI has recently become a huge part of the research effort in AI and machine learning. As a result, methodology papers have been published, with for example the work of \newcite{bender2018data}, for NLP data and systems, encouraging the community towards rich and explicit data statements. Figure~\ref{fig.1} shows the evolution of gender information availability in the last 10 years. We can see that this peek of interest is also present in our data, with more resources provided with gender information after 2017.

\section{Recommendations}

The social impact of big data and the ethical problems raised by NLP systems have already been discussed by previous work. \newcite{wilkinson2016fair} developed principles for scientific data management and stewardship, the FAIR Data Principles, based on four foundational data characteristics that are Findability, Accessibility, Interoperability and Reusability \cite{wilkinson2016fair}. In our case, findability and accessibility are taken into account by design, resources on OpenSLR being freely accessible. Interoperability and Reusability of data are however not yet achieved. Another attempt to integrate this discussion about data description within the NLP community has been made by \newcite{COUILLAULT14.424}, who proposed an Ethics and Big Data Charter, to help resources creators describe data from a legal and ethical point of view. \newcite{hovy2016social} highlighted the different social implications of NLP systems, such as \emph{exclusion}, \emph{overgeneralisation} and \emph{exposure} problems. More recently, work by \newcite{bender2018data} proposed the notion of data statement to ensure data transparency.

The common point of all these studies is that information is key. The FAIR Principles are a baseline to guarantee the reproducibility of scientific findings. We need data to be described exhaustively in order to acknowledge demographic bias that may exist within our corpora. As pointed out by \newcite{hovy2016social}, language is always situated and so are language resources. This demographic bias in itself will always exist, but by not mentioning it in the data description we might create tools and systems that will have negative impacts on society. The authors presented the notion of \emph{exclusion} as a demographic misrepresentation leading to exclusion of certain groups in the use of a technology, due to the fact that this technology fail to take them into account during its developing process. This directly relates to our work on ASR performance on women speech, and we can assume that this can be extended to other speaker characteristics, such as accent or age. To prevent such collateral consequences of NLP systems, \newcite{bender2018data} advocated the use of data statement, as a professional and research practice. We hope the present study will encourage researchers and resources creators to describe exhaustively their data sets, following the guidelines proposed by these authors.

\subsection{On the Importance of Meta-Data}
The first take-away of our survey is that obtaining an exhaustive description of the speakers within speech resources is not straightforward. This lack of meta-data is a problem in itself as it prevents guaranteeing the generalisability of systems or linguistics findings based on these corpora, as pointed out by \newcite{bender2018data}. As they rightly highlighted in their paper, the problem is also an ethical one as we have no way of controlling the existence of representation disparity in data. And this disparity may lead to bias in our systems. 

We observed that most of the speech resources available contain elicited speech and that on average, researchers are careful as to balance the speakers in terms of gender when crafting data. But this cannot be said about corpora containing non-elicited speech. And apart from Librispeech, we observed a general gender imbalance, which can lead to a performance decrease on female speech \cite{Garnerin:2019:GRF:3347449.3357480}. Speech time measurements are not consistent throughout our panel of resources and utterance counts are not reliable. We gathered the size of the corpora as well as the sampling rate in order to estimate the amount of speech time available, but variation in terms of precision, bit-rate, encoding and containers prevent us from reaching reliable results. Yet, speech time information enables us to know the quantity of data available for each category and this directly impacts the systems. This information is now given in papers such as the one describing the latest version of TEDLIUM,\footnote{However, as gender information was not provided with the release we used, we did not take it into account in our survey.} as this information is paramount for speaker adaptation.

\newcite{bender2018data} proposed to provide the following information alongside corpus releases: curation rationale, language variety, speaker demographic, annotator demographic, speech situation, text characteristics, recording quality and others. Information we can add to their recommendations relates to the duration of the data sets in hours or minutes, globally and per speaker and/or gender category. This could allow to quickly check the gender balance in terms of quantity of data available for each category, without relying on an unreliable notion of utterance. This descriptive work is of importance for the future corpora, but should also be made for the data sets already released as they are likely to be used again by the community.

\subsection{Transparency in Evaluation }
Word Error Rate (WER) is usually computed as the sum of the errors made on the test data set divided by the total number of words. But if such an evaluation allows for an easy comparison of the systems, it fails to acknowledge for their performance variations. 
In our survey, 13 of the 66 corpora had a paper describing the resources. When the paper reported ASR results, none of them reported gendered evaluation even if gender information about the data was provided. Reporting results for different categories is the most straightforward way to check for performance bias or overfitting behaviours. Providing data statements is a first step towards, but for an open and fair science, the next step should be to also take into account such information in the evaluation process. A recent work in this direction has been made by \newcite{mitchell2019model} who proposed to describe model performance in model cards, thus encouraging a transparent report of model results.

\section{Conclusion}

In our gender survey of the corpora available on the OpenSLR platform, we observe the following trends: parity is globally achieved on the whole, but interactions with other corpus characteristics reveal that gender misrepresentation needs more than just a number of speakers to be identified. In non-elicited data (meaning type of speech that would have existed without the creation of the corpus, such as TEDTalks or radio broadcast), we found that, except in Librispeech where gender balance is controlled, men are more represented than women. It also seems that most of the corpora aimed at developing TTS systems contain mostly female voices, maybe due to the stereotype associating female voice with caring activities. We also observe that gender description of data has been taken into account by the community, with an increased number of corpora provided with gender meta-data in the last two years. Our sample containing only 66 corpora, we acknowledge that our results cannot necessarily be extended to all language resources, however it allows us to open discussion about general corpus description practices, pointing out a lack of meta-data and to actualise the discourse around the social implications of NLP systems. We advocate for a more open science and technology by following guidelines such as the FAIR Data Principle or providing data statements, in order to ensure scientific generalisation and interoperability while preventing social harm.

\section{Acknowledgements}
This work was  partially supported by MIAI@Grenoble-Alpes (ANR-19-P3IA-0003).

\section{Copyrights}

The Language Resources and Evaluation Conference (LREC)
proceedings are published by the European Language Resources Association (ELRA).
They are available online from the conference website.

ELRA's policy is to acquire copyright for all LREC contributions. In assigning
your copyright, you are not forfeiting your right to use your contribution
elsewhere. This you may do without seeking permission and is subject only to
normal acknowledgement to the LREC proceedings. The LREC 2020 Proceedings are
licensed under CC-BY-NC, the Creative Commons Attribution-Non-Commercial 4.0
International License.

\section{Bibliographical References}\label{reference}

\bibliographystyle{lrec}
\bibliography{lrec2020W-xample-kc,references}

\section{Language Resource References}
\label{lr:ref}
\bibliographystylelanguageresource{lrec}
\bibliographylanguageresource{languageresource}

\end{document}

%% file: tables/info_prov.tex
\begin{table}
\centering
\begin{tabular}{llc}
\hline
\textbf{Gender info available}  &           & \textbf{Number of corpora}  \\
\hline
No              &           & 24 (36.4\%)           \\
\hline
Yes             & metadata  & 9 (13.6\%)            \\
                & indexed   & 28 (42.4\%)           \\
                & paper     & 5 (7.6\%)              \\
\hline
\textbf{Total}  &    -      & \textbf{66} \\
\hline
\end{tabular}
 \caption{Information availability on gender in OpenSLR corpora.}
 \label{tab:info_prov}
\end{table}

%% file: tables/prov_spk_dur.tex
\begin{table}
\centering
\begin{tabular}{lc}
\hline
\textbf{Gender info available}  &  \textbf{Number of corpora}  \\
\hline
Number of speakers              &     41                \\

Number of utterances            &    32                 \\

Speech time                     &     5                 \\
\hline
\textbf{Total number of corpora}& \textbf{42}                    \\
\hline
\end{tabular}
 \caption{Type of information provided in terms of gender alongside the 42 corpora containing gender information.}
 \label{tab:info_prov_gender}
\end{table}

%% file: tables/gender_elicited.tex
\begin{table}
\centering
\begin{tabular}{lccc}
\hline
\textbf{Type of speech}      & \textbf{\#corpora} & \textbf{\#F} & \textbf{\#M}  \\
\hline
Elicited            & 41        & 1782              & 1596          \\
                    &           & 52.8\%            & 47.2\%        \\
\hline
Non-elicited        & 5         & 1268              & 1426          \\
                    &           & 47.1\%            & 52.9\%        \\
\hline
Non-elicited        & 4         & 67                & 143           \\
(without Librispeech) &         & 31.9\%            & 68.1\%        \\
\hline
\end{tabular}
    \caption{Speaker gender distribution in data depending on the type of speech. \emph{NB: the two last lines refer to the non-elicited corpora, the only difference is that the last line does not take Librispeech into account.}}
    \label{tab:gender_rep_el}
\end{table}

%% file: tables/gender_lang_status.tex
\begin{table}
\centering
\begin{tabular}{lcccc}
\hline
\textbf{Language status} & \textbf{\#corpora} & \textbf{\#F}    & \textbf{\#M}  & \textbf{Total}\\
\hline
Low-resource    & 23        & 677       & 539       & 1216   \\
                &           & 55.7\%    & 44.3\%    & 100\% \\
\hline
High-resource   & 19        & 1105      & 1057      & 2162  \\
                &           & 51.1\%    & 48.9\%    & 100\%  \\
\hline
\end{tabular}
\caption{Speaker gender distribution in elicited corpora depending on language status.}
 \label{tab:gender_rep_stat}
\end{table}

%% file: tables/asr_tts_table.tex
\begin{table}
\centering
\begin{tabular}{lccc}
\hline
\textbf{Task}    & \textbf{\#corpora} & \textbf{\#F}   & \textbf{\#M}  \\
\hline
ASR     & 12    & 2523              & 2615          \\
        &       & 49.1\%            & 50.9\%        \\
\hline
TTS     & 10    & 124               & 70            \\
        &       & 63.9\%            & 36.1\%        \\
\hline
NA      & 25    & 403               & 337           \\
        &       & 54.5\%            & 45.5\%        \\
\hline
\end{tabular}
\caption{Speaker gender representation in data depending on the task. ASR stands for Automatic Speech Recognition, TTS stands for Text To Speech, and NA accounts for the corpora for which no task was explicitly cited.} 
    \label{tab:gender_rep_task}
\end{table}

%% file: tables/women_tts_table.tex
\begin{table}
\centering
\begin{tabular}{lcc}
\hline
\textbf                         & \textbf{F}        & \textbf{M}  \\
\hline
Number of speakers              & 591               & 551     \\
                                & 51.8\%            & 48.2\%     \\
\hline
Number of utterances            & 72,280            & 143,342     \\
                                & 33.5\%            & 66.5\%     \\
\hline
\end{tabular}
\caption{Number of speakers of each gender and number of utterances for each gender category within the subset of corpora providing utterance count by gender. \emph{N.B: two corpora provided utterance count by gender but no speaker count, so the number of speakers is only given as a trend.}}
 \label{tab:gender_rep_tts}
\end{table}